\documentclass[twocolumn,final,authoryear]{elsarticle}

\usepackage{framed,multirow}

\usepackage{amssymb}
\usepackage{latexsym}

\usepackage{graphicx}
\usepackage{subfigure}
\usepackage{threeparttable}
\usepackage{tabularx}

\usepackage{url}
\usepackage{xcolor}
\definecolor{newcolor}{rgb}{.8,.349,.1}
\DeclareGraphicsExtensions{.pdf,.png,.jpg}

\begin{document}

\ifpreprint
  \setcounter{page}{1}
\else
  \setcounter{page}{1}
\fi

\begin{frontmatter}

\title{Classifying Fonts and Calligraphy Styles Using Complex Wavelet Transform}

\author[1]{Alican Bozkurt\corref{cor1}} 

\ead{alicanb@gmail.com}
\author[2] {Pinar Duygulu}
\author[1]{A. Enis Cetin}

\address[1]{Department of Electrical and Electronics Engineering, Bilkent University, TR-06800 Bilkent, Ankara, Turkey}
\address[2]{Present address: Carnegie Mellon University, Department of Computer Science, 15213, Pittsburgh, PA, USA}
\cortext[cor1]{Corresponding author: Tel: +90-312-290 1477, 
}
\begin{abstract}
Recognizing fonts has become an important task in document analysis, due to the increasing number of available digital documents in different fonts and emphases. A generic font-recognition system independent of language, script and content is desirable for processing various types of documents. At the same time, categorizing calligraphy styles in handwritten manuscripts is important for palaeographic analysis, but has not been studied sufficiently in the literature. We address the font-recognition problem as analysis and categorization of textures. We extract features using complex wavelet transform and use support vector machines for classification. Extensive experimental evaluations on different datasets in four languages and comparisons with state-of-the-art studies  show that our proposed method achieves higher recognition accuracy while being computationally simpler. Furthermore, on a new dataset generated from Ottoman manuscripts, we show that the proposed method can also be used for categorizing Ottoman calligraphy with high accuracy.
\end{abstract}


\end{frontmatter}



\vspace{-0.5cm}
\section{Introduction}
\vspace{-0.2cm}
\label{sec:intro}
The term {\bf font} generally refers to a document's  typeface, such as {\tt  Arial}. Each font can have variations such as {\tt bold} or {\tt italic} to emphasize the text, which are called emphases. {\bf Font recognition}, which is the process of classifying different forms of letters, is an important issue in document analysis especially in multi-font documents~\citep{Slimane2013,Khosravi2010}. In addition to its advantages in capturing document layout, font recognition may also help to increase the performance of optical character recognition (OCR) systems by reducing the variability of shape and size of the characters. 

For printed fonts in languages that use the Latin alphabet, the main challenge is to recognize fonts in ``noisy'' documents, that is, those containing many artifacts. When we consider languages with cursive scripts, such as Arabic, the change in character shape with location (isolated, initial, medial or final) and dots (diacritics) above or below the letters cause further difficulties in character segmentation.

Handwritten documents add an extra component to the analysis because of  writing style. Classifying handwriting styles continues to be a challenging yet important problem for palaeographic analysis~\citep{Aiolli1999}. In recent studies of Hebrew, Chinese and Arabic calligraphy, researchers used characters as the basic elements to extract features~\citep{Yosef2007,Zhuang2009,Azmi2011}. However, these methods heavily rely on preprocessing steps and are prone to error.

Another example which we focus, the style of Ottoman calligraphy, is the artistic handwriting style of Ottoman Turkish. Different styles were used in different documents, such as books, letters, etc. ~\citep{Rado1983}.
Scholars around the world want to access efficiently and effectively to Ottoman archives, which contain millions of documents. Classifying Ottoman calligraphy styles would be an important step in categorizing large numbers of documents in archives, as it would assist further processing for retrieval, browsing and  transliteration.


The Ottoman alphabet is similar to Farsi and Arabic in the sense that it uses a right-to-left-cursive script. 
Hence, to recognise multi-font printed texts in Ottoman \citep{Ozturk-2000}, existing methods of Arabic and Farsi font recognition can be utilised. 
However, due to the late adoption of printing technology in the Ottoman Empire, a high percentage of documents are handwritten. Documents in Ottoman calligraphy are very challenging compared to their printed counterparts, with intra-class variances much higher than what is found in printed documents. Some documents are hundreds of years old and the non-optimal storage conditions of historic Ottoman archives have resulted in highly degraded manuscripts. So as not to damage the binding, books are scanned with their pages only partially open, introducing non-uniform lighting in images. 

We propose a simple but effective method for recognizing printed fonts independent of language or alphabet, and extend it to classifying handwritten calligraphy styles, particularly Ottoman manuscripts. We present a new method based on analyzing textural features extracted from text blocks, which therefore does not require complicated preprocessing steps such as connected component analysis or segmentation. While Gabor filters are commonly used in the literature for texture analysis, they are computationally costly~\citep{Khosravi2010}. Alternatively, we propose using complex wavelet transform (CWT), which is not only more efficient but also achieves better recognition rates compared to other methods.

Unlike most existing studies focusing on a single language, we experiment on many previously studied printed fonts in four languages: English, Farsi, Arabic and Chinese. Our method also yields high accuracy in categorizing Ottoman calligraphy styles on a newly generated dataset consisting of various samples of different handwritten Ottoman calligraphy styles.

\vspace{-0.5cm}
\section{Related Work}
\label{sec:relatedwork}
\vspace{-0.2cm}

One of the first systems of optical font recognition was \citep{Zramdini1998}, in which global typographical features were extracted and classified by a multivariate Bayesian classifier. The authors extracted eight global features from connected components.
They experimented on ten typefaces in seven sizes and four emphases in printed and scanned English documents. \cite{Ozturk-2001} proposed a cluster based approach for printed fonts and exploited recognition performance for quality analysis.

Feature extraction methods for font recognition in the literature are divided into two basic approaches: local and global. Local features, usually refer to the typographical information gained from parts of individual letters, and are utilised in~\citep{chaudhuri1998automatic,juan2005font}. Local feature extraction relies on character segmentation, requiring the documents to be noise free and scanned in high resolution. Global features refer to information extracted from entire words, lines or pages, and are mostly texture based  \citep{amin1998off,abuhaiba2004arabic,borji2007support,Khosravi2010,Slimane2013}.

Zhu et al.~\citep{Zhu2001} addressed the font recognition problem as a texture identification issue, and used multichannel Gabor filters to extract features. Prior to feature extraction, the authors normalized the documents to create uniform blocks of text. Experimental results were reported on computer-generated images with 24 Chinese (six typefaces and four emphases) and 32 English (eight typefaces and four emphases) fonts. Pepper and salt noise was added to generate artificial noise.
Gabor filters were also used in ~\citep{Ramanathan2009} for feature extraction, and support vector machines (SVM) for classification. Experiments were carried out on six typefaces with four emphases on English documents. Similarly, Ma and Doermann used Gabor filters not for font identification but for script identification at word level. The authors used three different classifiers, SVM, k-nearest neighbour and the Gaussian mixture model (GMM), to identify the scripts in four different bilingual dictionaries (Arabic-English, Chinese-English, Hindi-English, Korean-English). The method was also used to classify Arial and Times New Roman fonts~\citep{MaDoermann2003}.
\citeauthor{AvilesCruz2005} used high-order statistical moments to characterize the textures, and Bayes classifier for classification in~\citep{AvilesCruz2005}. Similar to~\citep{Zhu2001}, they experimented on Spanish documents digitally generated with eight fonts and four emphases. They also tested the effects of Gaussian random noise.

~\citep{Khosravi2010} approached the Farsi font recognition problem. Instead of using Gabor filter-based method, the authors proposed a gradient-based approach to reduce the computational complexity. They combined Sobel and Roberts operators to extract the gradients, and used AdaBoost for classification. In~\citep{SenobariKhosravi2012}, Sobel-Robert operator-based features were combined with wavelet-based features. Another method, based on matching interest points is proposed in \citep{Zahedi2011}.

For Arabic font recognition, Ben Moussa et al.~\citep{BenMoussa2008,BenMoussa2010} proposed a method based on fractal geometry
The authors generated a dataset consisting of printed documents in ten typefaces. \citep{Slimane2013} proposed a method for recognising fonts and sizes in Arabic word images at an ultra-low resolution. They use GMM to model the likelihoods of large numbers of features extracted from grey level and binary images. Bataineh et al. considered statistical analysis of edge pixel behavior in binary images  for feature extraction from Arabic calligraphic scripts in \citep{Bataineh2012}. They  experimented on Kufi, Diwani, Persian, Roqaa, Thuluth and Naskh styles.

For classifying calligraphy styles, \citep{Yosef2007} presented a writer-identification method based on extracting geometric parameters from three letters in Hebrew calligraphy documents, followed by dimension reduction.  \citep{Azmi2011} proposed using triangle blocks for classifying Arabic calligraphy. In this method, triangles are formed using tangent values and grey-level occurrence matrices extracted from individual characters. \citep{Zhuang2009} introduced a generative probabilistic model for automatically extracting a presentation in calligraphic style for Chinese calligraphy works. The authors created a latent style model based on the latent Dirichlet allocation model.
\vspace{-0.5cm}
\section{Proposed Method}
\label{sec:approach}
\vspace{-0.2cm}

We propose a new method for categorising writing styles, applicable to printed fonts and calligraphy styles. Given a text in a particular language, our goal is to classify the writing style as one of the known categories. Our method is robust to noise, computationally efficient, and extendable to handwritten documents. We consider documents as textures, and use CWT, which has the ability to capture directional features at various angles and scales in a computationally efficient manner. We describe the details of the proposed method in the following.

\vspace{-0.2cm}
\subsection{Preprocessing}
It is assumed that the input is a grey-level image of the text to be classified, and that empty margins around the text are cropped. Since the focus of this study is to classify the fonts, but not the layout extraction, text areas are cropped manually.

The proposed method has the ability to work on multi-font documents, and it can also be used for segmenting parts in different fonts. Moreover, our method is capable of detecting and discarding empty areas, which is achieved by block processing. A binarized document image is divided into blocks, and a block is marked as ``empty'' if the ratio of black pixels to white pixels is below a certain threshold, meaning that there is not sufficient text in that block. By properly choosing the block size, blocks can be arranged to include minimal heterogeneous text (in terms of font) as much as possible. Choosing block size is further discussed in Section \ref{sec:analysis}. We use Otsu's method~\citep{otsu} for binarization, which is observed to be effective for the documents we experiment with. Non-empty blocks are fed to a feature extraction process.

It is common to do a normalization to fix the font sizes before performing feature extraction. This process usually requires line and character length prediction, with projection profiles commonly used for this purpose. After the normalization step, most studies use  space filling to generate uniform text blocks \citep{AvilesCruz2005}. The normalization step was not required for our study because we generated artificial datasets with a fixed font size, and because Ottoman documents have relatively similar font sizes. 

\vspace{-0.2cm}
\subsection{Complex wavelet transform for feature extraction}
\label{sec:fextraction}
\begin{figure}
     \centering
     \begin{small}
     \begin{tabular}{cccccc}
{\includegraphics[width=0.08\linewidth]{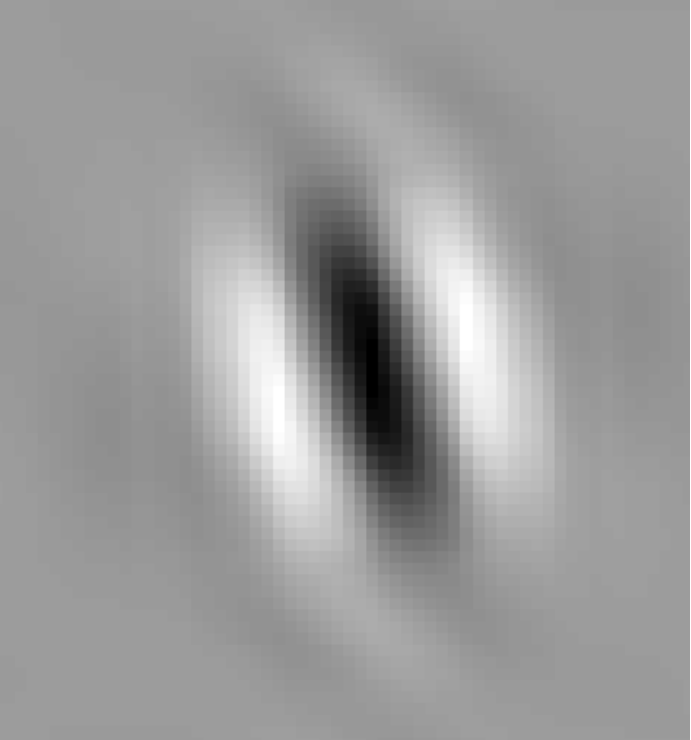}} &
{\includegraphics[width=0.08\linewidth]{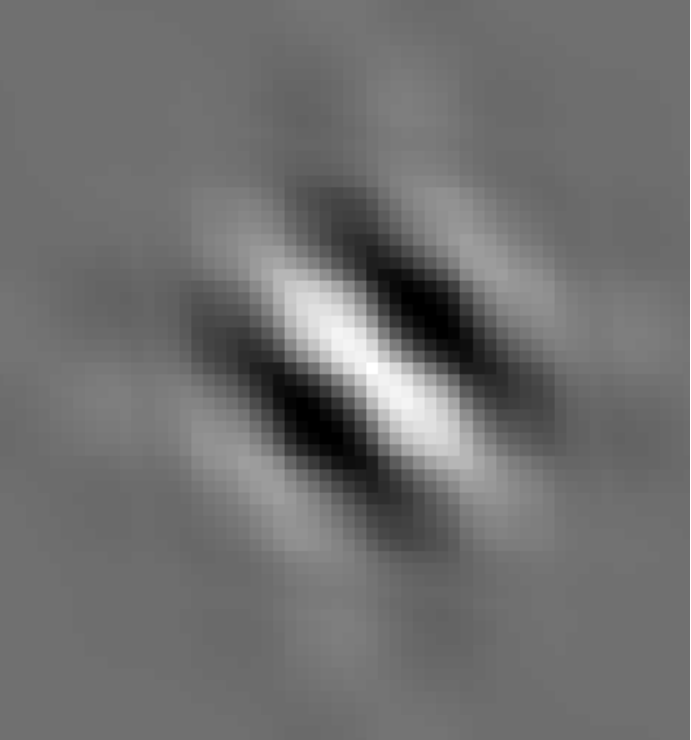}} &
{\includegraphics[width=0.08\linewidth]{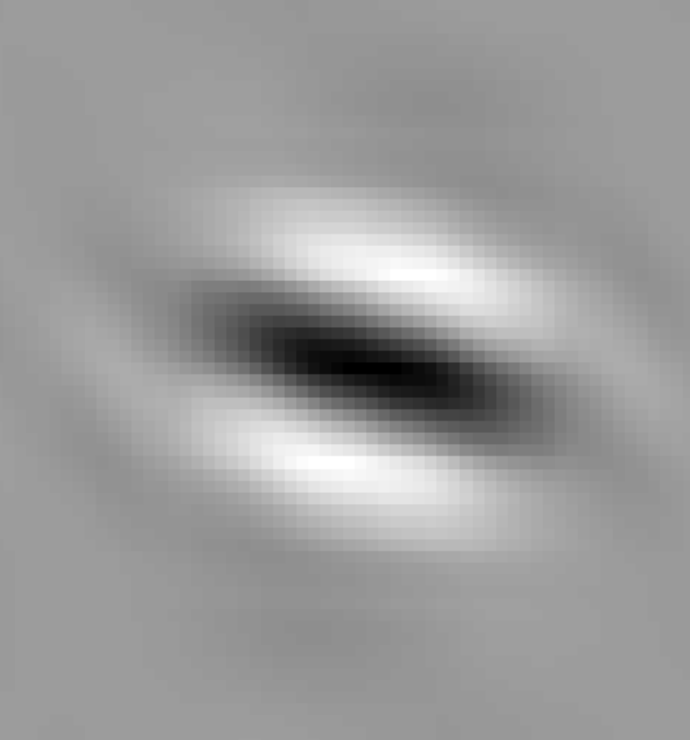}} &
{\includegraphics[width=0.08\linewidth]{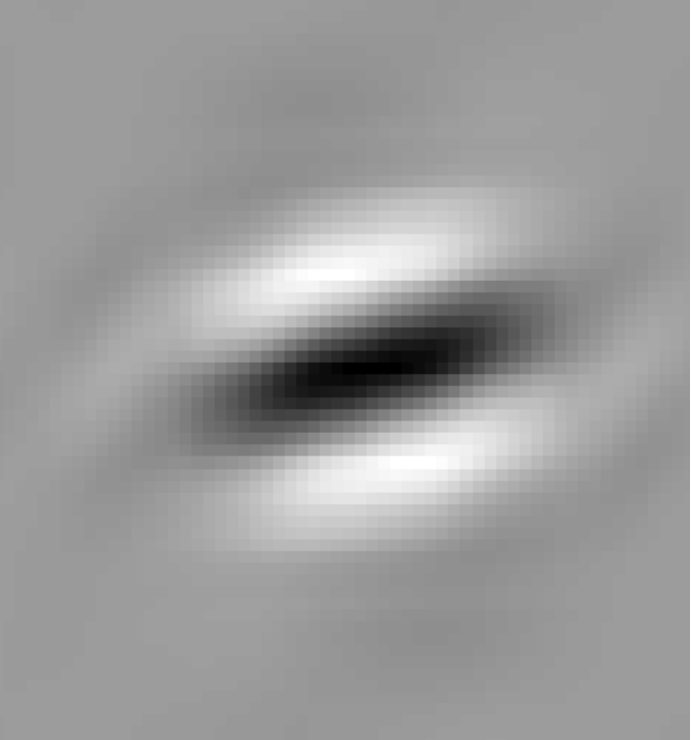}} &
{\includegraphics[width=0.08\linewidth]{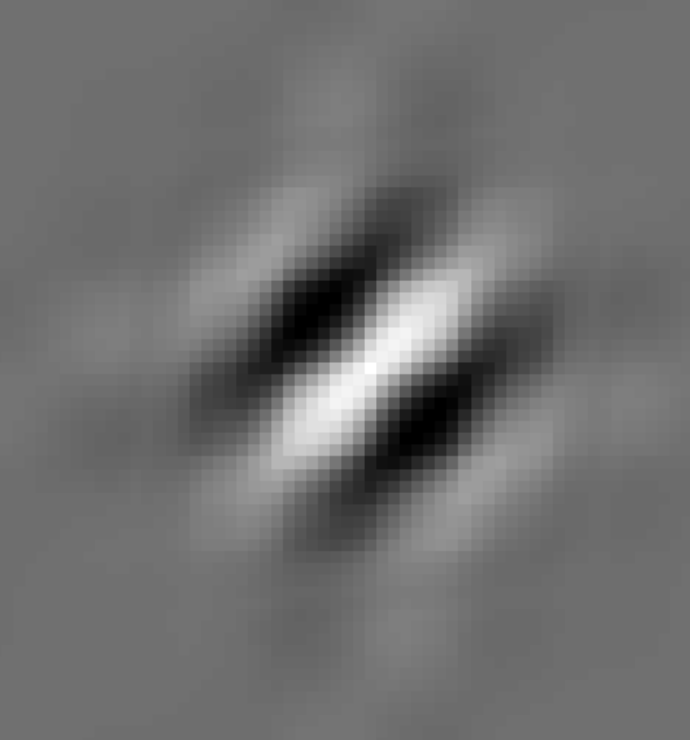}} &
{\includegraphics[width=0.08\linewidth]{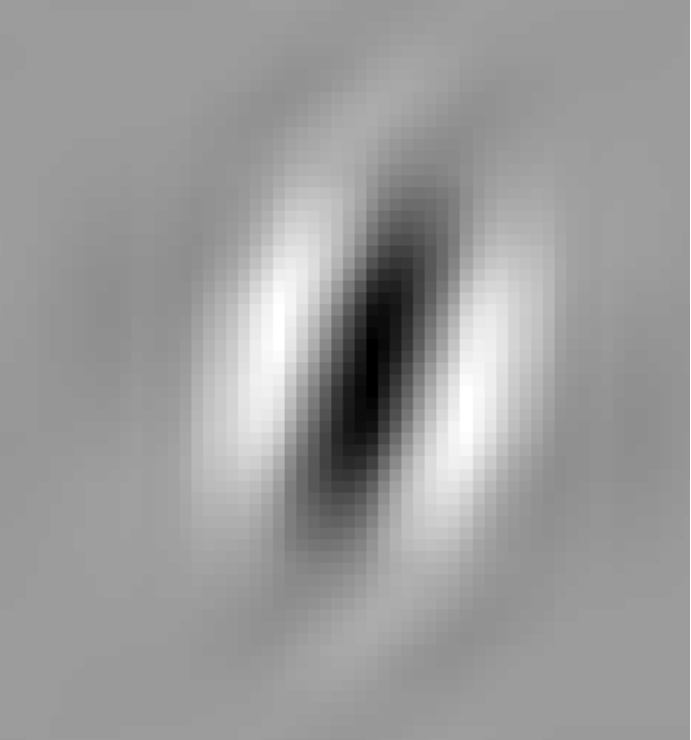}} \\
-75 & -45 & -15 & 15 & 45 & 75 \\
{\includegraphics[width=0.08\linewidth]{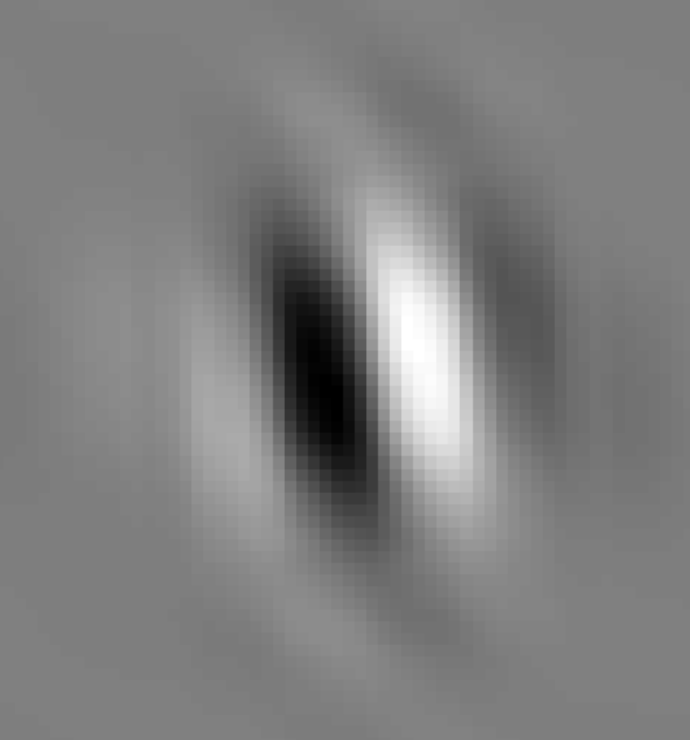}} &
{\includegraphics[width=0.08\linewidth]{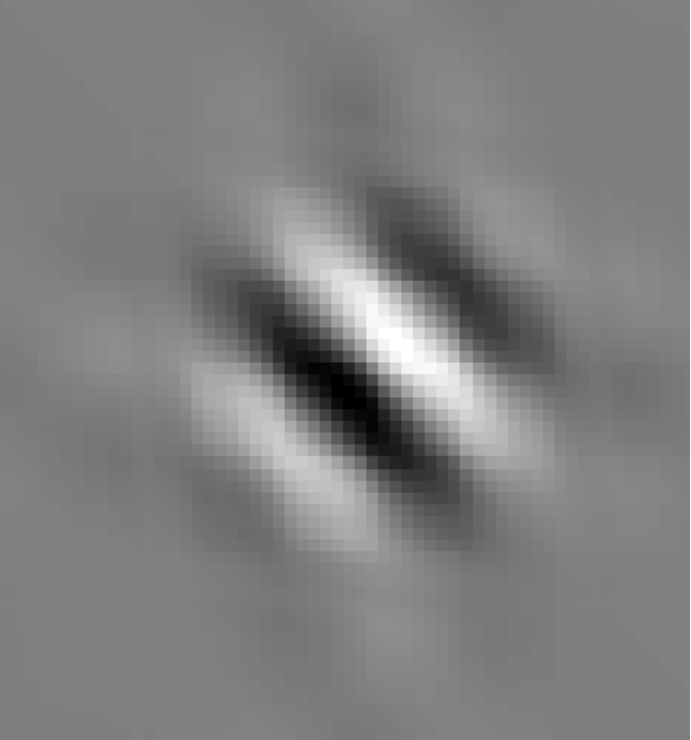}} &
{\includegraphics[width=0.08\linewidth]{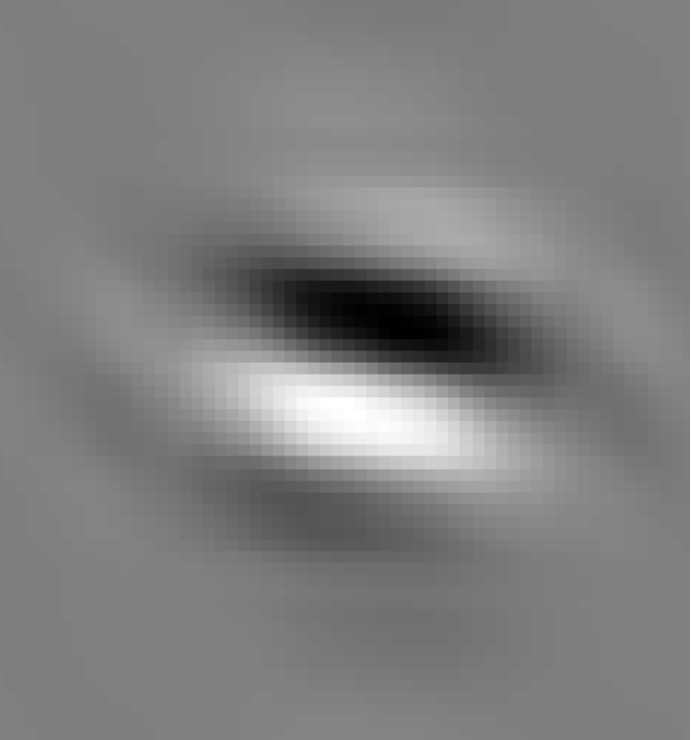}} &
{\includegraphics[width=0.08\linewidth]{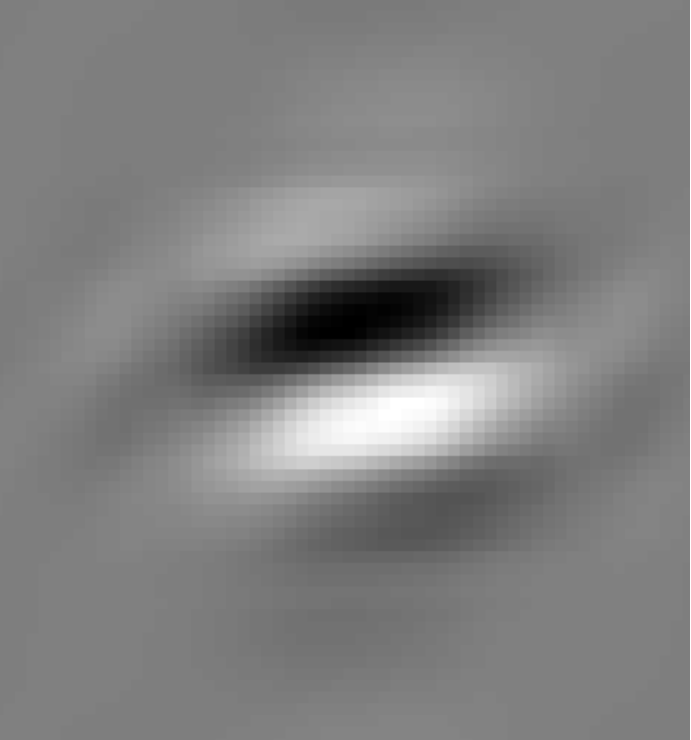}} &
{\includegraphics[width=0.08\linewidth]{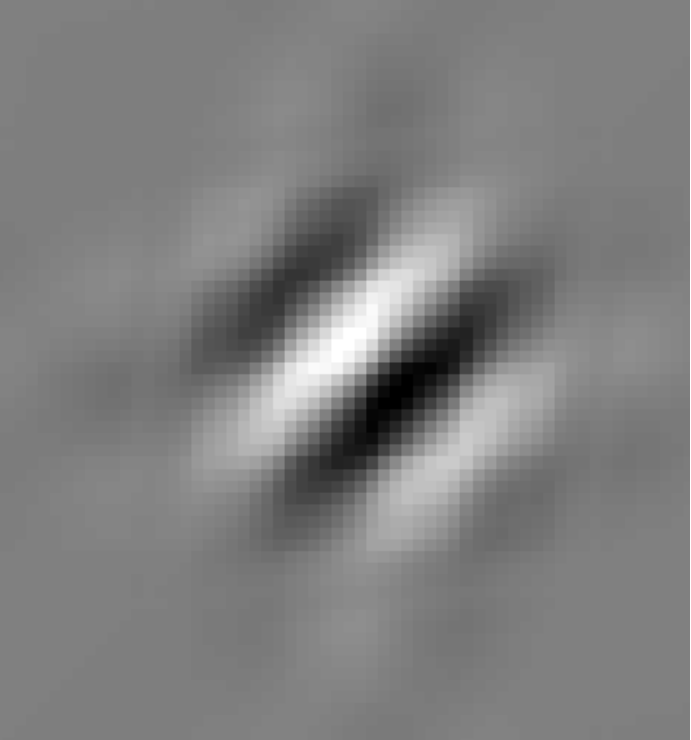}} &
{\includegraphics[width=0.08\linewidth]{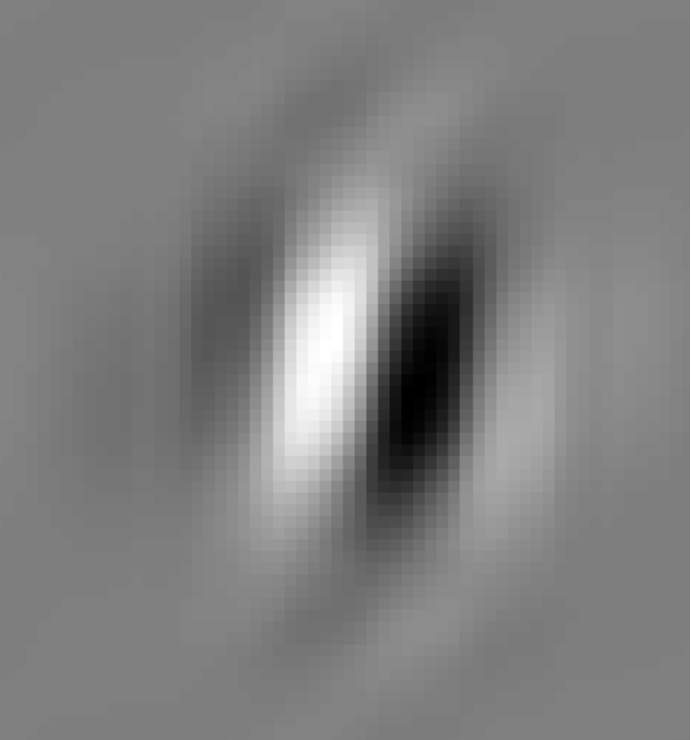}} \\
-75 & -45 & -15 & 15 & 45 & 75 \\
\end{tabular}
\end{small}
\vspace{-0.2cm}
\caption{2D Impulse responses of CWT at level 4 (top:real, bottom:imaginary).}
\vspace{-0.5cm}
\label{fig:impulse_responses}
\end{figure}

For feature extraction, we use CWT \citep{hill2000rotationally,celik2009multiscale,hatipoglu1999texture,portilla2000parametric}. Ordinary discrete wavelet transform (DWT) is not as reliable as CWT when modeling textural features, because the former is shift-variant and only sensitive to horizontal ($0^{\circ})$ and vertical $(90^{\circ})$ directions.
On the other hand, dual-tree complex wavelet transform (DT-CWT), proposed in \citep{kingsbury1998dual}, is almost shift-invariant, directionally selective at angles $\pm 15^{\circ}$, $\pm 45^{\circ}$ and  $\pm 75^{\circ}$, and has perfect reconstruction capability. It introduces minimal redundancy (4:1 for images) and has a computational complexity of $O(N)$ \citep{kingsbury1998dual}. 
There are many choices for wavelet filters and filter banks; in this work, we use Farras filters Farras \citep{farras2001nearly} and the six-tap filter designed by \citep{kingsbury2000dual} (see Table \ref{table:filters1}).  Note that $L_K^n,H_K^n,L_F^n$ and $H_F^n$ only show the real parts of the filters. Imaginary parts of the filters can be calculated by reversing and taking the negative of each respective filter \citep{kingsbury2000dual,farras2001nearly}.
A dual tree is constructed from $L_K^n,H_K^n,L_F^n$, and $H_F^n$. Using this tree it is possible to decompose the input image into directional sub-bands with CWT (see Figure~\ref{fig:impulse_responses}).  After a single stage of DT-CWT image decomposition, the image is decomposed into directional sub-bands with orientations of $\pm 15^{\circ},45^{\circ}$ and $75^{\circ}$.

\begin{table}
\caption{(L)ow and (H)igh pass (K)ingsbury and (F)arras coefficients.}
\label{table:filters1}
\resizebox{\columnwidth}{!}{
\begin{tabular}{|c|c|c|c|c|c|c|c|c|c|c|}
\hline
$L_K^n$ & 0 & -0.0884 & 0.0884 & 0.6959 & 0.6959 & 0.0884 & -0.0884 & 0.0112 & 0.0112  & \\
$H_K^n$ & 0.0112 & 0.0112 & -0.0884 & 0.0884 & 0.6959 & 0.6959 & 0.0884 & -0.0884 & 0 & 0  \\
\hline
$L_F^n$  & 0.0351 & 0 & -0.0883 & 0.2339 & 0.7603 & 0.5875 & 0 & -0.1143 & 0 & 0\\
$H_F^n$ & 0 & 0 & -0.1143 & 0 & 0.5875 & 0.7603 & 0.2339 & -0.0883 & 0 & 0.0351\\
\hline
\end{tabular}
}
\end{table}

Since DT-CWT produces output images with different sizes at each tree level due to decimation, and these sizes depend on the input image size, it is not feasible to use output images of DT-CWT directly. Instead, we use statistical features of outputs of the complex wavelet tree, that is, the first and the second moments (i.e., mean and variance), because they are computationally more efficient and more robust to noise than higher-order moments. In experimenting with several levels of the complex wavelet tree, we find that recognition rate does not increase noticeably after three level trees. Overall, our feature vector includes mean and variance values of 18 output images (six outputs per level of a three-level complex wavelet tree), resulting in a 36-element feature vector. We use the MATLAB implementation of DT-CWT given by \citep{ShihuaCai}.

\vspace{-0.2cm}
\subsection{Classification}
We use SVMs \citep{VapnikSVM} for classification. The radial basis function (RBF) is used as the kernel function. As discussed in \citep{hsu2003practical}, RBF non-linearly maps samples into a higher dimensional space, which makes it very efficient for problems with small numbers of features, as in our case. We used LIBSVM \citep{chang2011libsvm}. The parameters of SVM and the RBF kernel, $C$ and $\gamma$, are optimized by a grid search and cross-validation for each case. The search range  is  $[1-10^6]$ for $C$ and $[10^{-6}-1]$ for $\gamma$.

\vspace{-0.5cm}
\section{Experimental Results}
\label{sec:results}
\vspace{-0.2cm}
To compare our method with state-of-the-art studies, we use the available datasets provided by other studies and generate artificial datasets of our own. For English, we create lorem ipsum texts using Microsoft \textregistered Word 2010, with fonts used in \citep{AvilesCruz2005}. In a similar fashion, we create paragraphs with fonts used in \citep{Zhu2001} for Chinese, and with fonts used in \citep{Khosravi2010} for Farsi. For Arabic, we perform our experiments on a publicly available dataset, used in \citep{BenMoussa2010}. In addition, we construct a new dataset of Ottoman calligraphy by scanning pages from Ottoman documents written in different calligraphy styles.

Although there are differences between the Arabic, Farsi and Ottoman languages, they use the same alphabet, with small variations. We use different datasets for Farsi and Arabic  to compare our method with  state-of-the-art methods for Farsi \citep{Khosravi2010} and Arabic \citep{BenMoussa2008}.  For instance, we use ``Farsi texts'' in this section to indicate the texts created with fonts used in \citep{Khosravi2010}, not as a general term for all documents in Farsi. The Ottoman dataset is fundamentally different from the Arabic and Farsi datasets because the former is in a handwritten form, whereas the latter two are created digitally. The procedure is called ``calligraphy style recognition'' for the Ottoman dataset, and ``font recognition'' for the others to emphasize the difference.

In the following, we first present the experimental results for recognize printed fonts in each  dataset separately. Then, we show that our method is also capable of recognizing fonts in multi-font documents and that it can categorize fonts in multiple languages without a noticeable decrease in performance. We then present the results of the calligraphy style recognition. Finally, we provide a detailed analysis of our method.

\vspace{-0.2cm}
\subsection{Font Recognition}
This section presents the results for the English, Chinese, Farsi and Arabic documents with printed fonts. Descriptions of the datasets are followed by experimental evaluations of that dataset. The accuracy of the model for recognizing different fonts is computed over a 10-fold cross-validation. 

{\bf Dataset 1 - English Texts:} 
\label{sec:datasets}
To test the noise performance of our method, we constructed three different English datasets. The first dataset, called ``noise-free'', consists of saved pages of English lorem ipsum texts typed in eight  typefaces (Arial, Bookman, Courier, Century Gothic, Comic Sans MS, Impact, Modern, Times New Roman), and in four emphases (regular, italic, bold, bold-italic). We directly converted these to images without any modification. The term ``noise-free'' here means that no noise is introduced in generating or saving the texts, because text images are directly fed to the methods. This set is used for validating and comparing methods in an ideal case. 
We also created noisy versions of the same texts by printing and scanning the pages in 200 dpi, as done in \citep{AvilesCruz2005}, using a Gestetner MP 7500 printer/scanner. This process introduced a small amount of noise to the images, hence we called the dataset created from these pages ``low-noise''.
We constructed the third dataset by photocopying and scanning the texts 10 times in succession, which resulted in a clear degradation of image quality and introduced a large number of artefacts. This dataset is an approximation of a worn-out document, and called ``high-noise''.
The average signal-to-noise ratio (SNR) value is  8.2487 for the low-noise dataset and 7.4062 for the high-noise dataset.  
These values are calculated by registering each image in the noisy dataset to the corresponding image in the noise-free dataset. The noise-free image is then subtracted from the registered image to extract the noise image. 

We compared our proposed method with the methods in~\citep{AvilesCruz2005} and in~\citep{Ramanathan2009} on the three datasets described above. We present the results in Table~\ref{tab:english}, with overall accuracy calculated over the four emphases for each font.
In ``noise-free'', all methods classify all fonts perfectly. As noise is introduced, our method begins to outperform the other methods in almost all fonts. Gabor filters' performance \citep{Ramanathan2009} is close to CWT because both methods employ directional filters at various angles. In the``high-noise'' dataset, because \citep{AvilesCruz2005} uses horizontal and vertical projections in the preprocessing stage, the noise and artifacts in the images prevent the method from segmenting the words properly and the method cannot provide any result. Our method, suffers only a $1,8\%$ decrease in average accuracy.

\begin{table}[h]
\vspace{-0.5cm}
\caption{Recognition rates ($\%$) of the proposed method, and comparisons with the methods of (1) \citep{AvilesCruz2005} and (2) \citep{Ramanathan2009} on English datasets. The highest accuracies are shown in bold. }
\centering
\resizebox{0.9\columnwidth}{!}{
\begin{threeparttable}
{\begin{tabular}{r|ccc|ccc|ccc}
    \hline
    \multirow{2}[4]{*}{Font\tnote{1}} & \multicolumn{3}{c}{Noise-free} & \multicolumn{3}{c}{Low-noise} & \multicolumn{3}{c}{High-noise} \\
    \cline{2-10}
          & Ours & (1) & (2)
          & Ours & (1) & (2)
          & Ours & (1) & (2) \\
\hline
    A & \textbf{100} & \textbf{100} & \textbf{100} & 96.90 & 81.80 & \textbf{100} & \textbf{98.44} & \textbf{-} & 91.70 \\
    B & \textbf{100} & \textbf{100} & \textbf{100} & \textbf{100} & 87.00 & \textbf{100} & \textbf{98.40} & \textbf{-} & 88.90 \\
    CG & \textbf{100} & \textbf{100} & \textbf{100} & \textbf{98.50} & 69.80 & 97.20 & 92.20 & \textbf{-} & \textbf{94.40} \\
    CS & \textbf{100} & \textbf{100} & \textbf{100} & \textbf{100} & 75.50  & \textbf{100} & \textbf{100} & \textbf{-} & 97.20 \\
    CN & \textbf{100} & \textbf{100} & \textbf{100} & \textbf{100} & 96.30 & \textbf{100} & \textbf{100} & \textbf{-} & 94.40 \\
    I & \textbf{100} & \textbf{100} & \textbf{100} & \textbf{100} & 99.00 & \textbf{100} & \textbf{100} & \textbf{-} & 94.40 \\
    CM & \textbf{100} & \textbf{100} & \textbf{100} & \textbf{100} & 97.00 & \textbf{100} & \textbf{98.40} & \textbf{-} & 88.90 \\
    T & \textbf{100} & \textbf{100} & \textbf{100} & \textbf{100} & 91.00 & \textbf{100} & 98.44 & \textbf{-} & \textbf{100} \\
\hline
Mean & \textbf{100} & \textbf{100} & \textbf{100} & 99.40 & 87.20 & \textbf{99.70} & \textbf{98.20} & - & 93.80 \\
    \hline
    \end{tabular}}
\begin{tablenotes}
		\footnotesize
       \item[1]{{\bf A}rial,{\bf B}ookman,{\bf C}entury {\bf G}othic,{\bf C}omic {\bf S}ans MS,{\bf C}ourier {\bf N}ew,  {\bf I}mpact,{\bf C}omputer {\bf M}odern,{\bf T}imes New Roman}
     \end{tablenotes}
  \end{threeparttable}}
  \label{tab:english}%
\end{table}%


{\bf Dataset 2 - Chinese Texts:}
Since the proposed method uses textural information, which is not dependent on a specific language, there is no limitation in the language or the alphabet selection regarding the method applied. We show this advantage of our method by testing it on Chinese texts. We used four different emphases (regular, italic, bold, bold-italic) for six different fonts used in \citep{Zhu2001}: SongTi, KaiTi, HeiTi, FangSong, LiShu and YouYuan. This set can be considered the Chinese equivalent of noise-free English set.

We compare the proposed method to the Chinese font recognition methods described in \citep{Zhu2001} and \citep{Yang2006}. Gabor features are used in \citep{Zhu2001} and characters' stroke features are used in ~\citep{Yang2006}. Recognition rates of the methods for each font and style are presented in Table~\ref{tab:cf_chinese}. An average performance of 97.16\% was reported in \citep{Yang2006} for the six fonts, and a 98.58\% average performance was obtained in~\citep{Zhu2001}. Our method has the highest overall recognition accuracy 98.81\%.

\begin{table}
\vspace{-0.5cm}
  \centering
  \caption{Recognition accuracy of the proposed method, and (1) \citep{Yang2006}'s and (2) \citep{Zhu2001}'s method for the Chinese dataset. The highest accuracies are shown in bold.}
  \resizebox{0.8\columnwidth}{!}{
    \begin{tabular}{r|ccc|ccc|ccc}
    \hline
          & \multicolumn{3}{c}{SongTi} & \multicolumn{3}{c}{KaiTi} & \multicolumn{3}{c}{HeiTi} \\
    \hline
		& Ours & (1) & (2) & Ours & (1)  & (2) & Ours & (1)  & (2)  \\
\hline
    regular & \textbf{100} & 98.20  & \textbf{100} & \textbf{100} & \textbf{100} & 98.40 & 92.86 & 93.20  & \textbf{100} \\
    bold  & \textbf{100} & 93.20  & \textbf{100} & 95.24 & 91.70  & \textbf{98.40} & \textbf{100} & \textbf{100} & \textbf{100} \\
    italic & \textbf{100} & \textbf{100} & 97.60  & 95.24 & \textbf{100} & 89.60  & \textbf{100} & 96.50  & \textbf{100} \\
    bold italic & \textbf{100} & \textbf{100} & 99.20  & \textbf{100} & 93.30  & \textbf{100} & \textbf{100} & \textbf{100} & \textbf{100} \\
mean: & \textbf{100} &	97.85	& 99.20 &	\textbf{97.62} &	96.25 &	96.60 &	98.21 &	97.42 &	\textbf{100}\\
\hline
& \multicolumn{3}{c}{FangSong} & \multicolumn{3}{c}{LiShu} & \multicolumn{3}{c}{YouYuan} \\
\hline
& Ours & (1) & (2) & Ours & (1) & (2) & Ours & (1) & (2) \\
\hline
regular & \textbf{100} & \textbf{100} & 92.80  & \textbf{100} & \textbf{100} & \textbf{100} & \textbf{100} & \textbf{100} & 99.60 \\
    bold  & 95.24 & 92.90  & \textbf{100} & \textbf{100} & \textbf{100} & \textbf{100} & \textbf{100} & 96.50  & 99.60 \\
    italic & \textbf{100} & 93,30  & 94.00    & 92.86 & \textbf{100} & \textbf{100} & \textbf{100} & \textbf{100} & \textbf{100} \\
    bold italic & \textbf{100} & 91.70  & 96.80  & \textbf{100} & 96.50  & \textbf{100} & \textbf{100} & 94.90  & \textbf{100} \\
mean: & \textbf{98.81} &	94.47	 & 95.90 &	98.21 &	99.12 &	\textbf{100} &	\textbf{100} &	97.85 &	99,80\\
    \hline
    \end{tabular}}%
  \label{tab:cf_chinese}%
\end{table}%

{\bf Dataset 3 - Farsi Texts:}
The Ottoman alphabet is similar in nature to the Farsi alphabet. To compare the performance of the proposed method against \citep{Khosravi2010}, we replicated the dataset used in~\citep{Khosravi2010}, which consists of scanned pages of Farsi lorem ipsum paragraphs written in four different emphases (regular, italic, bold, bold-italic) in ten different fonts: Homa, Lotus, Mitra, Nazanin, Tahoma, Times New Roman, Titr, Traffic, Yaghut, and Zar. Only regular and italic are used for Titr, because bold emphasis is not available for that font. 

Table~\ref{tab:cf_farsi} presents recognition accuracies of the proposed method for Farsi texts and comparisons with \citep{Khosravi2010} and \citep{SenobariKhosravi2012}. Overall, our method performs better than the others.

\begin{table}
  \centering
  \caption{Recognition rates ($\%$) of the proposed method and comparisons with \citep{Khosravi2010} and \citep{SenobariKhosravi2012} for Farsi texts.}
  \resizebox{\columnwidth}{!}{
    \begin{tabular}{rccc}
    \hline
    \multicolumn{1}{c}{\multirow{2}[0]{*}{\textbf{Font}}} & \multicolumn{3}{c}{\textbf{Recognition Rates ($\%$)}}\\
    \cline{2-4}
    \multicolumn{1}{c}{} & Proposed & \citep{Khosravi2010}
    & \citep{SenobariKhosravi2012}\\
	\hline
    Lotus & \textbf{92.2}  & \textbf{92.2} & 90.7 \\
    Mitra & \textbf{95.3}  & 93,4 & 93.7 \\
    Nazanin & 90.6  & 85.2 & \textbf{92.0} \\
    Traffic & \textbf{98.4}  & 97,6 & 95.9 \\
    Yaghut & 96.9  & 97.6 & \textbf{98.5} \\
    Zar   & \textbf{92.2}  & 87.4 & 90.9 \\
    Homa  & \textbf{100}   & 99.2 & 99.8 \\
    Titr  & \textbf{100}   & 95.2 & 97.0 \\
    Tahoma & \textbf{100}   & 96.6 & 98.3\\
    Times & 98.4  & 97.2 & \textbf{98.8} \\
    \hline
    Mean & \textbf{96.41} & 94.16 & 95.56 \\
    \hline
    \end{tabular}
   }
   \label{tab:cf_farsi}%
\end{table}%

{\bf Dataset 4 - Arabic Texts:}
The ALPH-REGIM database, a dataset for printed Arabic scripts, is provided by~\citep{BenMoussa2008}. The dataset consists of text snippets of various fonts, sizes, and lengths. We use ten typefaces which are also used in~\citep{BenMoussa2008}: Ahsa, Andalus, Arabictransparant, Badr, Buryidah, Dammam, Hada, Kharj, Koufi and Naskh.
 
We compare our method with the work of Ben Moussa et al. ~\citep{BenMoussa2010} for the ALPH-REGIM dataset. Since this dataset contains images with various sizes, block size must be chosen accordingly. The smallest image in the dataset (in terms of height) is a $100\times 1322$ image containing two lines. Therefore, we choose a block size of $96\times 160$, to efficiently sample every image in the dataset. For six out of ten fonts, our method results in much better performances than the other methods (see Table~\ref{tab:arabic}). Our method also performs better when mean accuracy is considered.

\begin{table}
  \centering
  \caption{Recognition accuracy of the proposed method and the method of~\citep{BenMoussa2010} on Arabic text}
\resizebox{0.9\columnwidth}{!}{
  \begin{tabular}{rcc}
    \hline
  & Proposed & \multirow{2}[0]{*}{\citep{BenMoussa2010}} \\
  & ($96\times 160$ sample size) & \\
\hline
 Ahsa  & \textbf{99.63} & 94.00 \\
    Andalus & \textbf{98.77} & 94.00 \\
    Arabictransparant & \textbf{99.82} & 92.00 \\
    Badr  & 99.44 & \textbf{100} \\
    Buryidah & 98.30 & \textbf{100} \\
    Dammam & 99.95 & \textbf{100} \\
    Hada  & 90.39 & \textbf{100} \\
    Kharj & \textbf{90.35} & 88.00 \\
    Koufi & \textbf{99.35} & 98.00 \\
    Naskh & \textbf{98.57} & 98.00 \\
\hline
    Mean & \textbf{97.46} & 96.40 \\

    \hline
 \end{tabular}}
\label{tab:arabic}%
\end{table}%

\begin{figure}
     \centering
\begin{tabular}{ll}
\subfigure[Ground Truth]{\includegraphics[height=4.5 cm, width = 0.45\linewidth]{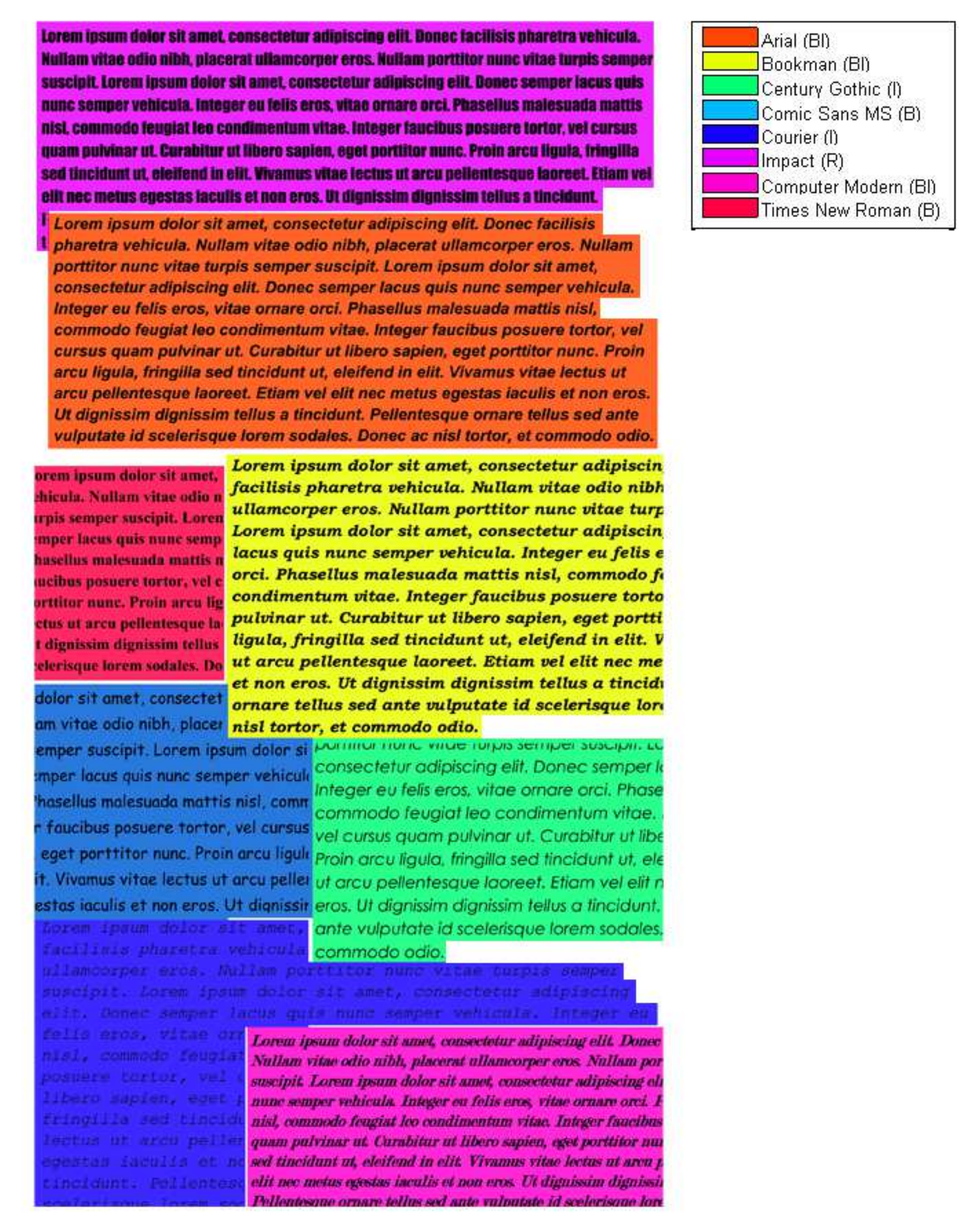}} &
\subfigure[Results]{\includegraphics[height=4.5 cm, width = 0.45\linewidth]{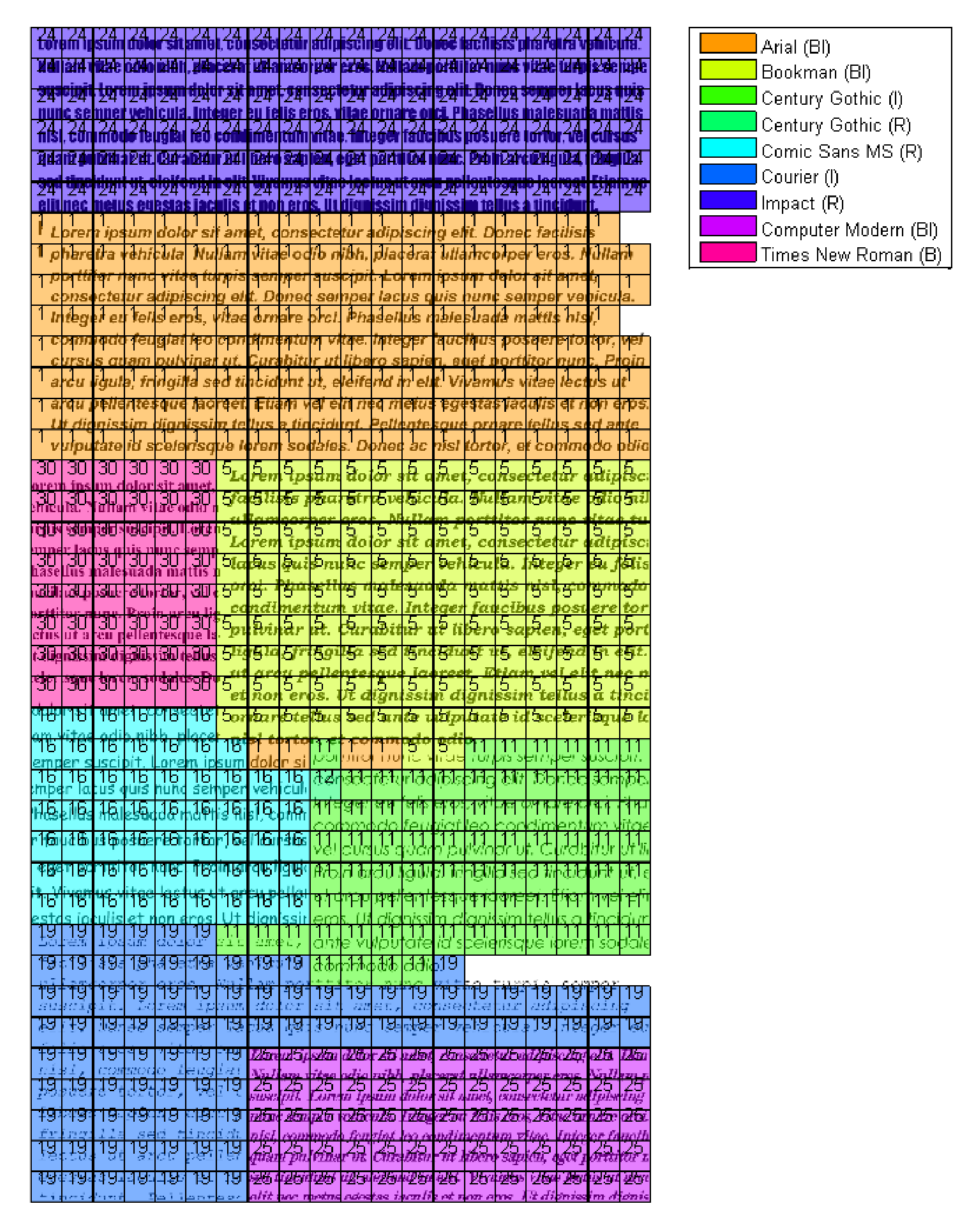}}
\end{tabular}
\caption{(a) A collage of fonts is generated to test the effectiveness of our method on multi-font documents.  Ground truth and estimated regions using our method are shown in (a) and (b), respectively.}
\label{fig:collage}
\end{figure}

{\bf Font recognition in documents with multiple fonts:}
To show that our method can also handle documents with multiple fonts, samples from the ``noise-free English'' dataset with $96 \times 96$ sized blocks were used to classify a collage of texts written in different fonts (see Figure~\ref{fig:collage}). 

{\bf Font recognition in the combined font dataset:}
As final test of font recognition, we considered all four datasets combined, resulting in a large dataset with 104 classes. We choose a block size of $96 \times 160$.
The average recognition accuracy of our method for each dataset, for both the single dataset case and the combined dataset case, are presented in Table~\ref{tab:combined}. The results show that the fonts of different alphabets do not overlap with each other in the proposed feature space and can be classified with an SVM without a noticeable decrease in performance.

\begin{table}
  \centering
  \caption{Average recognition accuracy of the proposed method for each dataset compared with  the combined dataset. Block size is $96 \times 160$.}
\resizebox{0.6\columnwidth}{!}{
\begin{tabular}{rcc}
\hline
Fonts & In Single Dataset & In the Combined Dataset \\
\hline
Farsi fonts & 87.23 & 86.88 \\
Arabic fonts & 96.86 & 96.75 \\
English fonts & 96.78 & 96.62 \\
Chinese fonts & 91.34 & 91.34 \\
\hline
\end{tabular}%
}
\label{tab:combined}%
\end{table}%



\vspace{-0.2cm}
\subsection{Calligraphy Style Recognition}
\begin{figure}
\subfigure[Divani]{\includegraphics[width=0.18\linewidth]{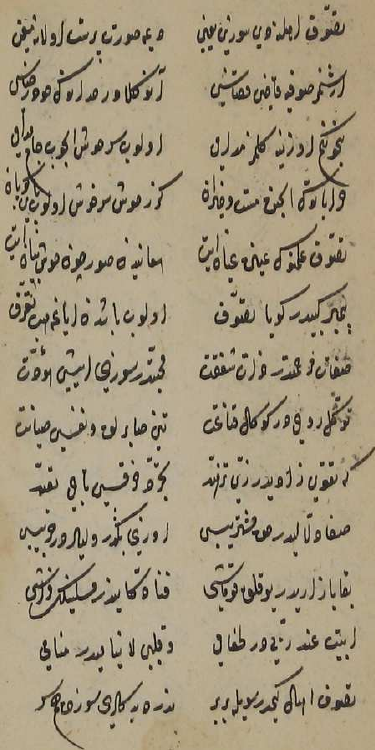}}
\subfigure[Matbu]{\includegraphics[width=0.18\linewidth]{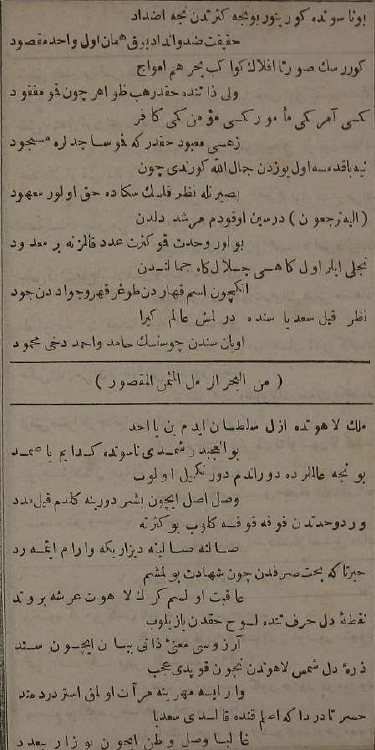}}
\subfigure[Nesih]{\includegraphics[width=0.18\linewidth]{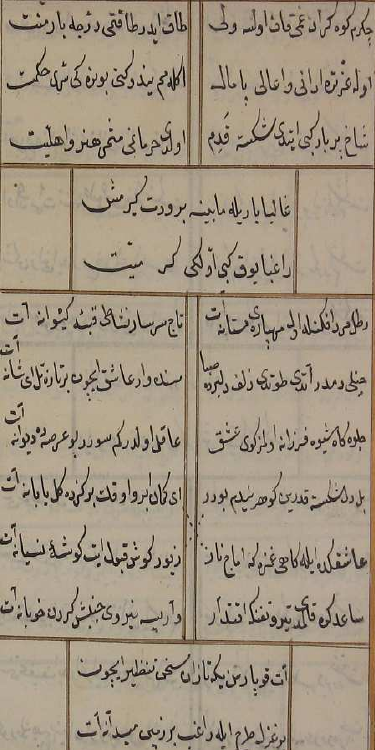}}
\subfigure[Rika]{\includegraphics[width=0.18\linewidth]{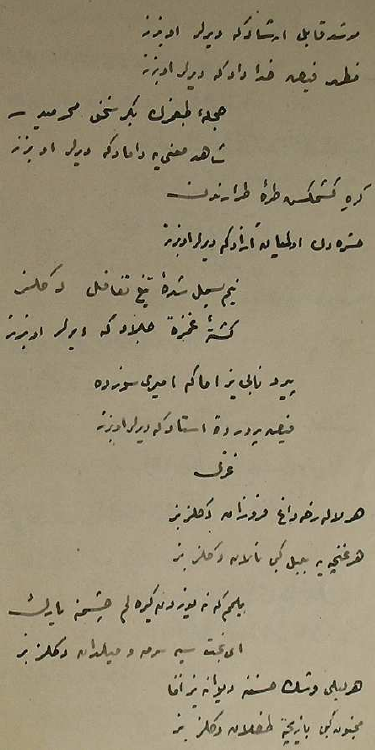}}
\subfigure[Talik]{\includegraphics[width=0.18\linewidth]{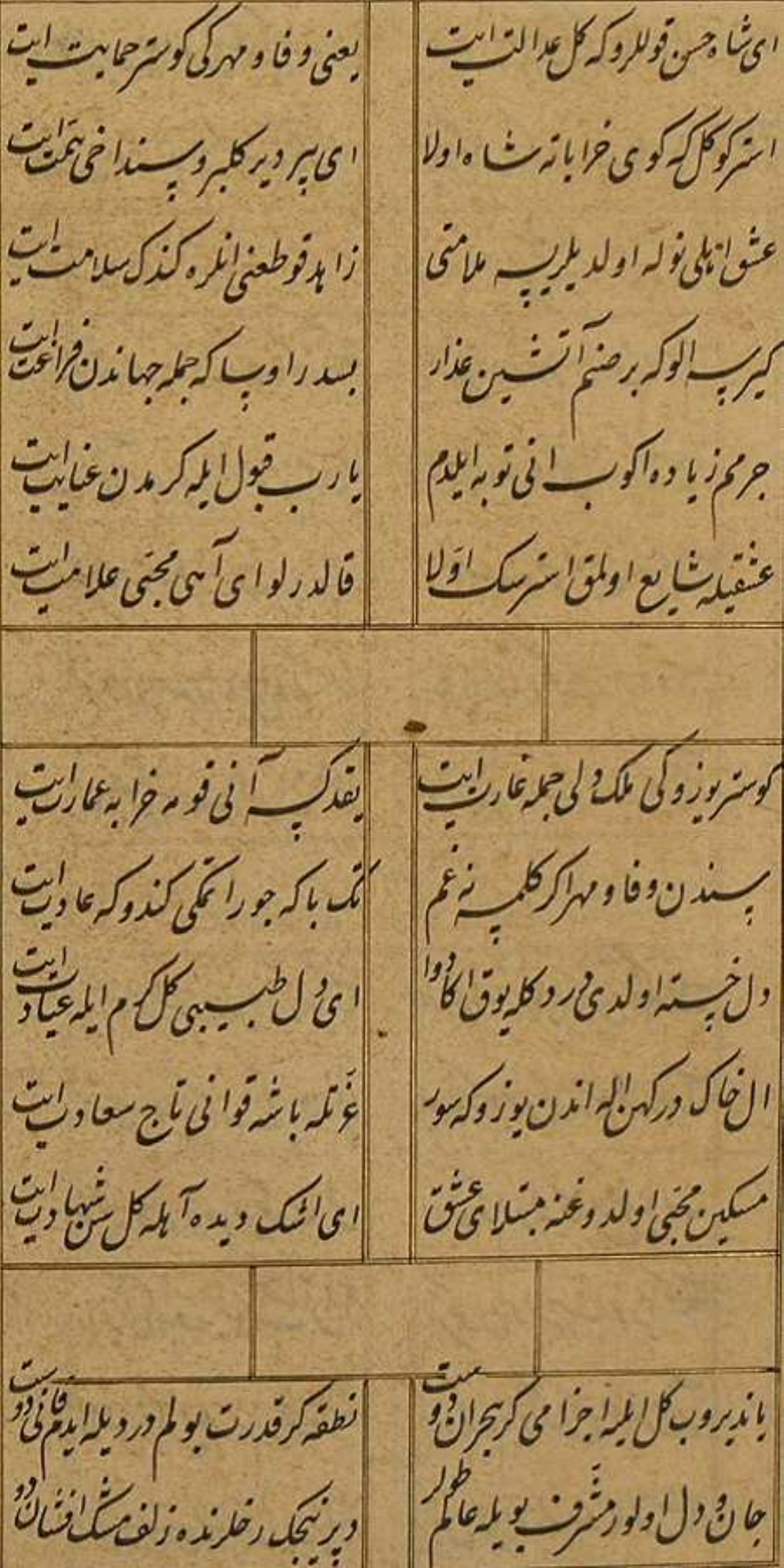}}
\caption{Example page images of different styles of Ottoman calligraphy.}
\label{fig:ottoman_fonts}
\vspace{-0.5cm}
\end{figure}

We show that the proposed method is capable of categorising calligraphy styles as well as printed forms through our experiments on a dataset generated from handwritten Ottoman manuscripts.
To the best of our knowledge, automatic classification of Ottoman calligraphy has not been studied. We created a new dataset from documents written in Ottoman calligraphy by scanning 30 pages in five different styles: {\tt divani, matbu, nesih, rika}, and {\tt talik}. Example documents from this dataset are presented in Figure~\ref{fig:ottoman_fonts}.
Ottoman documents have some characteristic page layouts that, together with calligraphy styles, is usually specific to a form of manuscript. These distinct layouts help with high accuracy discrimination. To normalise this layout difference, we created an additional set uniform in terms of layout. The areas that contain text are stitched together to eliminate empty spaces in the document. 


We performed two tests to classify calligraphy styles in Ottoman documents. In the first test, we used unmodified documents and extracted features from the entire image to make use of the page layout style. In the second test, we extracted blocks (as done in the font recognition steps) from the dataset images to test the effect of writing style alone.
Table~\ref{tab:cf_ottoman} summarizes the performance of the proposed method in the task of categorizing Ottoman calligraphy styles. Although there are large intra-class variations due to different handwriting characteristics, our method classifies the calligraphy styles almost perfectly.
In the second test, we choose a block size of  $96 \times 160$. The confusion matrix for the cropped dataset is given in Table~\ref{tab:cf_ottoman2}. The overall recognition accuracy is $96.01\%$, and the results indicate that, although the layout has a positive effect on accuracy, the proposed method can classify calligraphy styles with high accuracy using only texture information. 
\begin{table}
  \centering
  \caption{Confusion Matrix (recognition percentages) of the proposed method on unedited Ottoman texts.}
\resizebox{0.8\columnwidth}{!}{
    \begin{tabular}{r c c c c c}
	\hline
 \multicolumn{1}{c}{\multirow{2}[4]{*}{True Style}} & \multicolumn{5}{c}{Estimated Style} \\
\cline{2-6}
    \multicolumn{1}{c}{} & Divani & Matbu & Nesih & Rika & Talik\\
\hline
Divani & 98.95    & 0.00     & 0.00     & 1.05     & 0.00 \\
    Matbu & 0.00     & 100    & 0.00     & 0.00     & 0.00 \\
    Nesih & 0.00     & 0.00     & 100    & 0.00     & 0.00 \\
    Rika  & 0.00     & 0.00     & 0.00     & 100    & 0.00 \\
    Talik & 0.00     & 0.00     & 0.00     & 1.05     & 98.95 \\

    \hline
    \end{tabular} }
  \label{tab:cf_ottoman}%
\end{table}%

\begin{table}
  \centering
  \caption{Confusion Matrix (recognition percentages) of the proposed method on cropped Ottoman texts.}
\resizebox{0.8\columnwidth}{!}{
    \begin{tabular}{r c c c c c}
	\hline
 \multicolumn{1}{c}{\multirow{2}[4]{*}{True Style}} & \multicolumn{5}{c}{Estimated Style} \\
\cline{2-6}
    \multicolumn{1}{c}{} & Divani & Matbu & Nesih & Rika & Talik\\
\hline
Divani & 99.30   & 0.00     & 0.70     & 0.00     & 0.00 \\
    Matbu & 0.00     & 98.68    & 1.32     & 0.00     & 0.00 \\
    Nesih & 3.26     & 2.17     & 94.57    & 0.00     & 0.00 \\
    Rika  & 0.00     & 0.00     & 0.00     & 87.50    & 12.50 \\
    Talik & 0.00     & 0.00     & 0.00     & 0.00     & 100.00 \\

    \hline
    \end{tabular} }
  \label{tab:cf_ottoman2}%
\end{table}%

\vspace{-0.2cm}
\subsection{Analysis of the proposed method}
\label{sec:analysis}

{\bf Comparison of CWT and DWT:}
To demonstrate the superiority of CWT over DWT in the task of font recognition, the features are extracted from the English ``noise-free'' texts 
using both CWT and DWT.  The SVMs are trained and cross-validated using these features, and the SVM parameters are optimized for each case. The confusion matrices over all fonts and styles for DWT are presented in Table~\ref{tab:cf_dwt}.
``Correct'' and ``Wrong'' in Table~\ref{tab:cf_dwt} 
indicate the percentage of correctly and incorrectly classified samples, respectively. For example, the classifier using DWT outputs as features classifies 6.25\% of the samples' fonts correctly but classifies the style incorrectly (e.g. classifying Arial bold as Arial italic). DWT fail to differentiate between emphases of a font, especially bold/bold-italic and regular/italic, due to the lack of directional selectivity. CWT, on the other hand, perfectly discriminate among fonts and styles.
\begin{table}
  \vspace{-0.2cm}
  \centering
  \caption{Confusion matrix of classifier using DWT as features. Note that with CWT all the examples are correctly classified with 100\% accuracy.}
\resizebox{0.8\columnwidth}{!}{
    \begin{tabular}{rrr}
    \hline
          & Correct style & Wrong style \\
    \hline
    Correct font & 84.18 & 6.25 \\
    Wrong font & 5.27 & 4.30 \\
    \hline
    \end{tabular}}
  \label{tab:cf_dwt}%
\end{table}%

\begin{figure}
     \centering
  \begin{tabular}{ccc}
{\includegraphics[width=.1\hsize]{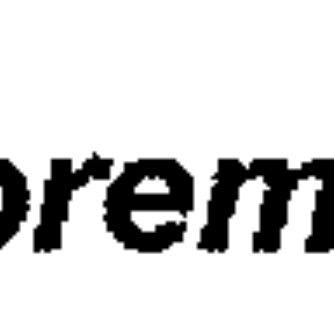}} &
{\includegraphics[width=.1\hsize]{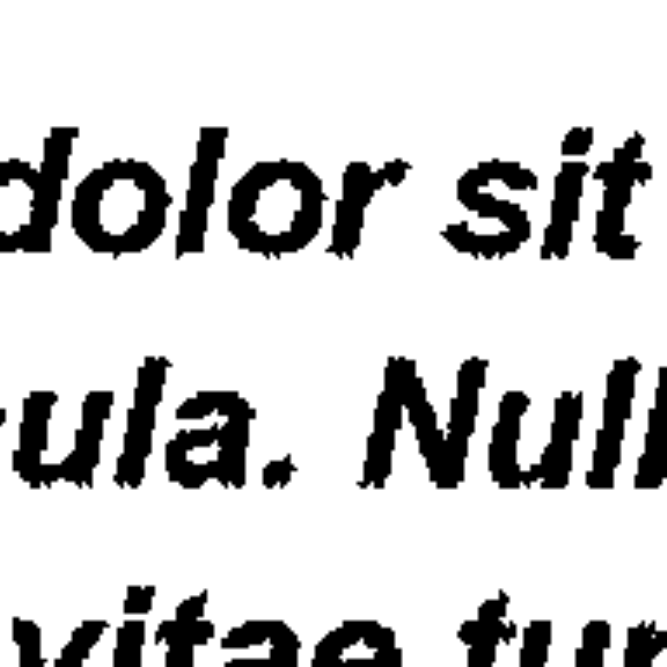}} & 
{\includegraphics[width=.1\hsize]{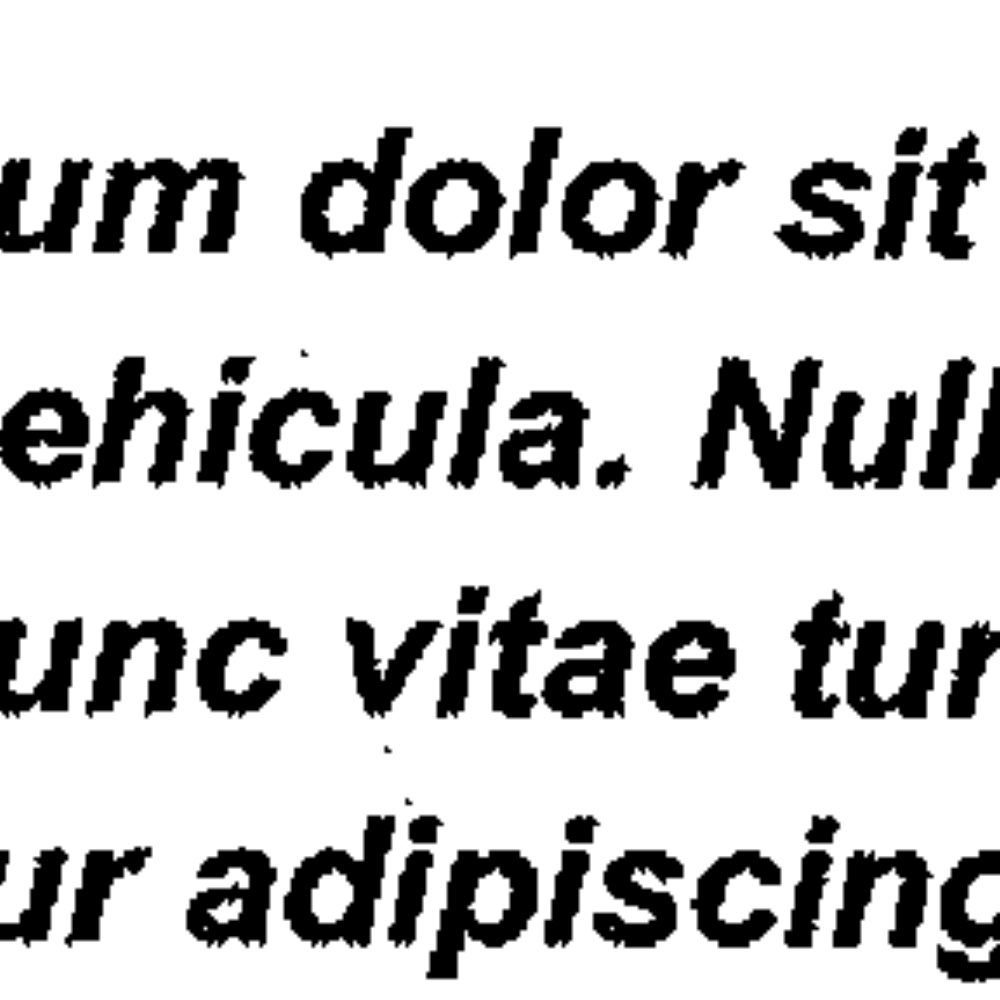}} \\
{\includegraphics[width=.3\hsize]{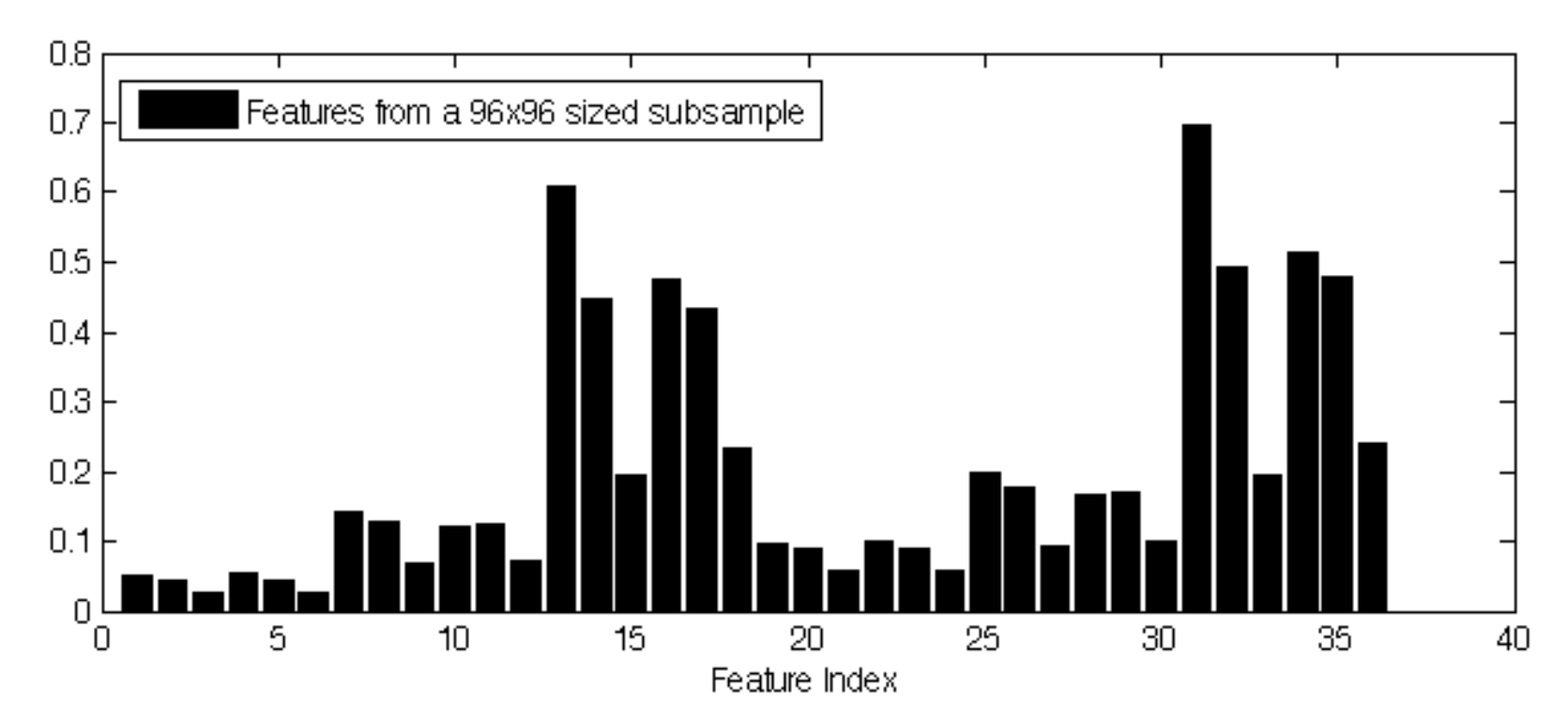}} &
{\includegraphics[width=.3\hsize]{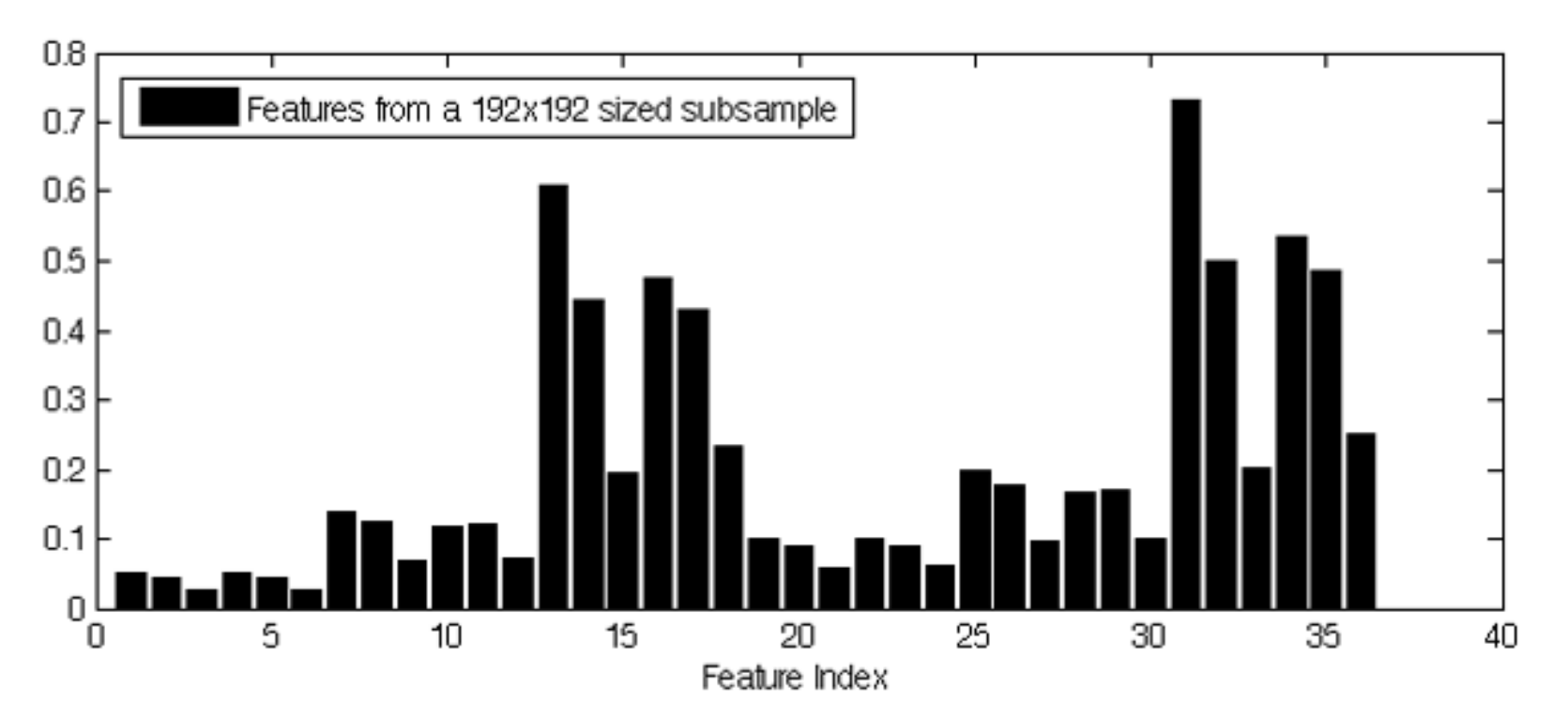}} &
{\includegraphics[width=.3\hsize]{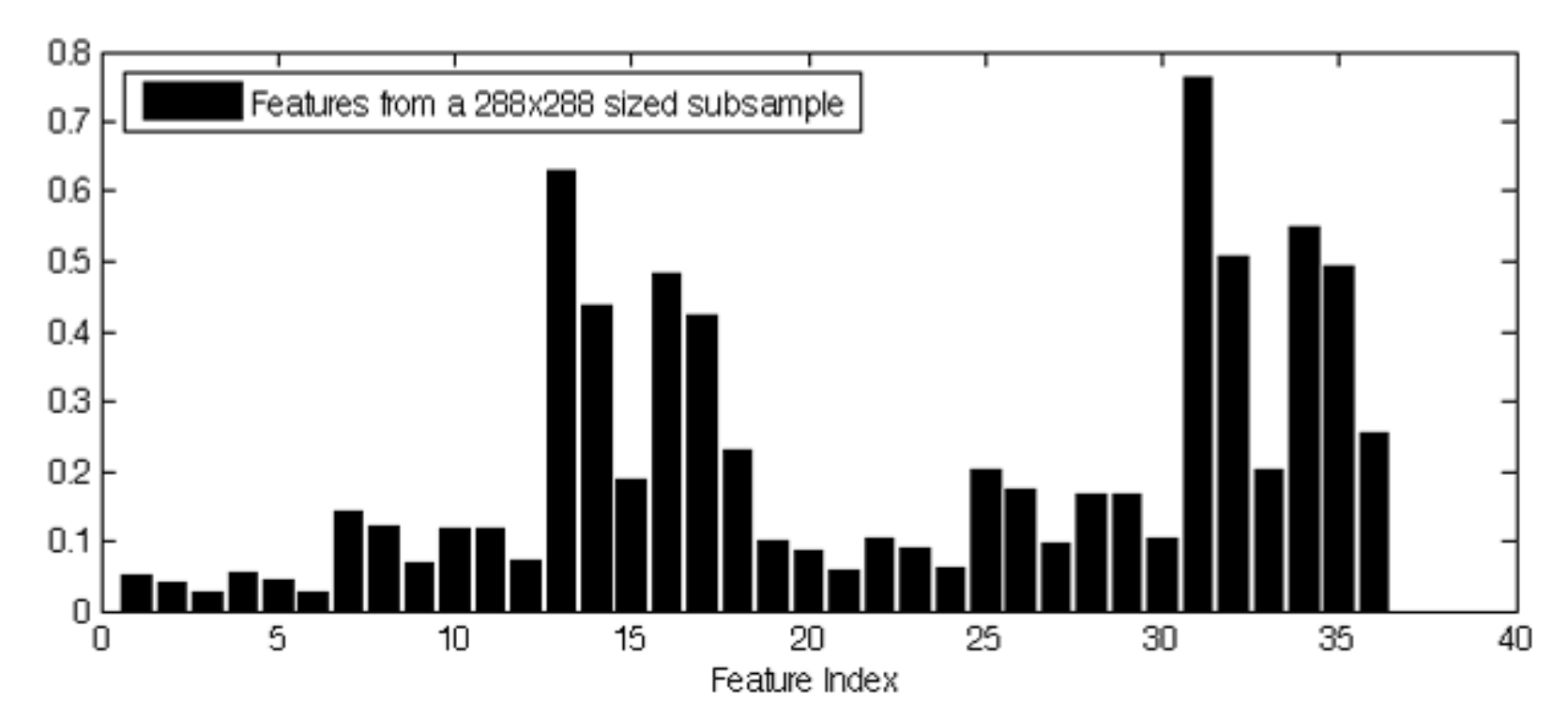}} \\
\end{tabular}
\vspace{-0.2cm}
\caption{Example image blocks (top) and average features of Arial bold (bottom), for block sizes (96x96), (192x192) and (288x288) respectively.}
\label{fig:sizetest}
\end{figure}

{\bf Choosing block size:}
\label{sec:blocksize}
Since decisions are generated per block, it is desirable that block size is as small as possible. However, the height of the blocks should be at least larger than the height of the characters in the sample, because smaller blocks would contain only parts of the characters and would not capture all characteristics of a given font.
Recall that we use 36 dimensional features corresponding to the mean and standard deviations of the absolute values of the outputs of the CWT. Statistical features allow a degree of freedom once that lower bound is passed. These features capture a font-style pair's characteristics very similarly, regardless of its block size. Figure~\ref{fig:sizetest} shows the features extracted from three different sizes of Arial bold blocks. As shown in Figure~\ref{fig:sizetest} , features of a $96 \times 96$ block are very similar to features of a $192 \times 192$ or $288 \times 288$ blocks. Therefore, a classifier trained with $96 \times 96$ blocks can easily classify $192 \times 192$ or $288 \times 288$  blocks.
To demonstrate, we performed several tests on English ``low-noise'' dataset.
A classifier was trained with $96\times 96$ blocks, and was used to classify $96\times 96$, $144\times 144$, $192\times 192$, $240\times 240$, $288\times 288$ and $336\times 336$ blocks. The results are presented in Table~\ref{tab:sizetest_96}.

\begin{table}
 \caption{Recognition rates of a classifier trained with features extracted from $96\times 96$ blocks.}
\resizebox{\columnwidth}{!}{
    \begin{tabular}{rrrrrrrr}
    \hline
    \multicolumn{1}{c}{\multirow{2}[0]{*}{Font}} & \multicolumn{1}{c}{\multirow{2}[0]{*}{Emphasis}} & \multicolumn{6}{c}{Block Size} \\
    \cline{3-8}
    \multicolumn{1}{c}{} & \multicolumn{1}{c}{} & $96\times 96$ & $144\times 144$ & $192\times 192$ & $240\times 240$ & $288\times 288$ & $336\times 336$ \\
    \hline
    Correct & Correct & 93.14 & 95.98 & 98.78 & 99.48 & 100 & 100 \\
    Correct & Wrong & 1.76  & 1.55  & 0.41  & 0.00  & 0.00  & 0.00 \\
    Wrong & Correct & 3.52  & 1.99  & 0.65  & 0.52  & 0.00  & 0.00 \\
    Wrong & Wrong & 1.57  & 0.49  & 0.16  & 0.00  & 0.00  & 0.00 \\
    \hline
    \end{tabular}}%
\label{tab:sizetest_96}%
\end{table}%

{\bf Computational Complexity:}
We compared the efficiency of our system with the other methods in terms of complexity. All tests were done in MATLAB \textregistered 2011b on a 32 bit Windows 7-installed PC with an Intel i7 1.6 GHz CPU and 4 GB RAM. Our feature extraction stage is faster in MATLAB than with Gabor, and SRF  implementations summarized in Table~\ref{tab:speed}. Our method has a lower complexity compared to other methods because the DT-CWT algorithm requires $O(N)$ multiplications and additions to $N$ input samples \citep{kingsbury1998dual}.

\begin{table}
\caption{Time to extract each feature from $128\times 128$ and $256\times 256$ sample.}
\resizebox{\linewidth}{!}{
  \begin{threeparttable}
    \begin{tabular}{rcrr}
    \hline
		\multicolumn{1}{c}{\multirow{2}[0]{*}{Feature}} & 	
\multicolumn{1}{c}{\multirow{2}[0]{*}{Implementation}} & 	
\multicolumn{2}{c}{Required Time per Sample (ms)} \\
\cline{3-4}
\multicolumn{2}{c}{} & $128\times 128$ & $256\times 256$ \\
\hline
    DT-CWT & MATLAB\tnote{a} \citep{ShihuaCai} & 4.40 & 10.40 \\
    \multicolumn{1}{r}{\multirow{2}[0]{*}{ SRF \citep{Khosravi2010}}} & MATLAB   &  9.40 & 13,70 \\
    \multicolumn{1}{c}{} & C & 3.78\tnote{b} & - \\
    Skewness \& Kurtosis \citep{AvilesCruz2005} & MATLAB & 8.60 & 39.30 \\
    Gabor \citep{Ramanathan2009} & MATLAB \citep{petkov2008gabor} & 29.30 & 100.70 \\
    \hline
    \end{tabular}%
	\begin{tablenotes}
		\footnotesize
       \item{[a]with precompiled C kernels, [b] value taken from \citep{Khosravi2010}}
     \end{tablenotes}
  \end{threeparttable}}
  \label{tab:speed}%
\vspace{-0.5cm}
\end{table}%

\vspace{-0.5cm}
\section{Conclusion}
\label{sec:conclusion}
\vspace{-0.3cm}
We present a novel and computationally efficient method for language and script-independent font and calligraphy style recognition. The mean and standard deviation of the absolute values of DT-CWT are used as features. SVM with an RBF kernel are trained for categorization. We compare the proposed method with  state-of-the-art studies on English, Farsi, Arabic and Chinese datasets. Experimental results indicate that our method outperforms the respective methods for all datasets. Our CWT-based features are computationally the most efficient among MATLAB implementations of the other feature-extraction methods. We also experimentally show that the proposed features are capable of capturing and discriminating among different font and calligraphic styles with high accuracy.

We show that texture-based image processing of handwritten Ottoman calligraphic documents is feasible. Calligraphic styles can be accurately determined by CWT-based image features and SVM. Since the style of an Ottoman document is an indicator of the type of document, calligraphic style estimation is an important step for automatic processing of the millions of Ottoman documents in archives.
It is possible to automate the entire Ottoman style-recognition method by incorporating region segmentation as a stage of pre-processing.  Image region segmentation can remove blank regions of a document, which  will not only speed up the entire font-recognition system but increase its accuracy.

In this article, we used our own Ottoman dataset, comprised of 60 documents. We are planning to enlarge our dataset by including other resources from the Internet and by scanning more Ottoman documents, with the aim of preparing a public database for automated Ottoman document processing.

We select the texture analysis block size as a multiple of the size of a single character. In general, each block contains several characters and character portions. It is also possible to automate  block size selection according to the size of a character by automatically detecting character sizes. We can also process a given document in overlapping blocks, which will increase the computational cost but  improve recognition results.

More complex preprocessing can be further investigated, to better accommodate real-life textures. Using such processing and the proposed features, it is possible to create a fast and accurate font detector that can be used on any document. One user-defined parameter in the proposed method is block size. Although flexibility in choosing this parameter is demonstrated, this it can also be automatized to efficiently enclose texts in a document. A future work could design an adaptive block size selection method. Although it was not necessary in our experiments, morphological or statistical filtering could be applied to SVM output to eliminate isolated block errors.



\bibliographystyle{model2-names}
\bibliography{Bozkurt-PRL_FS}

\end{document}